\documentclass{article} 
\usepackage{tccml_iclr2025_conference,times}

\usepackage{hyperref}
\usepackage{url}
\usepackage{booktabs}
\usepackage[retainorgcmds]{IEEEtrantools}
\usepackage{mathrsfs}
\usepackage{graphicx} 
\usepackage{amssymb}
\usepackage{bm}
\usepackage{dsfont}
\graphicspath{{figure/}}
\usepackage{algorithm}
\usepackage{algpseudocode}
\usepackage{color,soul}
\usepackage{mathtools}
\usepackage{multirow}
\usepackage{caption}
\usepackage{subcaption}
\usepackage{wrapfig}

\usepackage{enumitem}

\title{Generalizable Implicit Neural Representations via Parameterized Latent Dynamics for Baroclinic Ocean Forecasting}

\author{
Guang Zhao, Xihaier Luo, Seungjun Lee, Yihui Ren, Shinjae Yoo \\
Brookhaven National Laboratory, USA \\
\texttt{\{gzhao, xluo, slee19, yren, sjyoo\}@bnl.gov} \\
\And
Luke Van Roekel, Balu Nadiga \\
Los Alamos National Laboratory, USA \\
\texttt{\{lvanroekel, balu\}@lanl.gov} \\
\And
Sri Hari Krishna Narayanan, Yixuan Sun \\
Argonne National Laboratory, USA \\
\texttt{\{snarayan, yixuan.sun\}@anl.gov} \\
\And
Wei Xu \\
Brookhaven National Laboratory, USA \\
\texttt{xuw@bnl.gov}
}

\iclrfinalcopy 
\begin{document}

\maketitle
\begin{abstract}
  Mesoscale ocean dynamics play a critical role in climate systems, governing heat transport, hurricane genesis, and drought patterns. However, simulating these processes at high resolution remains computationally prohibitive due to their nonlinear, multiscale nature and vast spatiotemporal domains.  Implicit neural representations (INRs) reduce the computational costs as resolution-independent surrogates but fail in many-query scenarios (inverse modeling) requiring rapid evaluations across diverse parameters. We present \textbf{PINROD}, a novel framework combining dynamics-aware implicit neural representations with parametrized neural ordinary differential equations to address these limitations. By integrating parametric dependencies into latent dynamics, our method efficiently captures nonlinear oceanic behavior across varying boundary conditions and physical parameters. Experiments on ocean mesoscale activity data show superior accuracy over existing baselines and improved computational efficiency compared to standard numerical simulations. 
\end{abstract}

\section{Introduction}

Mesoscale ocean dynamics is important for climate modeling, as they directly influence global phenomena such as heat transport, hurricane formation, and drought prediction. These dynamics are governed by nonlinear physical processes across vast spatiotemporal scales, necessitating high-fidelity simulations for accurate understanding and prediction~\cite{deser2010sea,vallis2017atmospheric,luo2022bayesian}. However, the computational cost of high-resolution simulations of oceanic processes is prohibitive, particularly in many-query scenarios (e.g., parameter estimation) essential for handling the inherent uncertainties in initial conditions, boundary conditions, and model parameters~\cite{edwards2005uncertainties,almeida2010cooperative}.

Surrogate models, which approximate the input-output mapping of complex models, offer a computationally efficient alternative. While traditional reduced-order models like POD~\cite{berkooz1993proper} and DMD~\cite{schmid2010dynamic} are common, they often fail to capture nonlinear interactions in turbulent fluids. Recently, machine learning-based surrogates, particularly deep learning models, have shown promise in modeling spatiotemporal oceanic processes~\cite{berlinghieri2023gaussian,johnson2023oceanbench,lu2024oxygenerator}. Implicit neural representations (INRs) are particularly appealing due to their ability to learn continuous representations of spatial and temporal fields~\cite{xie2022neural,luo2024continuous}, crucial for representing the continuous nature of ocean dynamics.

Unlike grid-based models, INRs use coordinate-based neural networks, enabling extrapolation and continuous solutions. However, existing INR methods struggle with the demands of many-query scenarios in ocean modeling.  They typically require training separate networks for each set of conditions (initial, boundary, or model parameters), creating a significant computational bottleneck~\cite{lee2021meta,kim2023generalizable}. This hinders their practical application where rapid predictions across diverse scenarios are needed.

We introduce \textbf{P}arameterized \textbf{I}mplicit \textbf{N}eural \textbf{R}epresentations for \textbf{O}cean \textbf{D}ynamics (\textbf{PINROD}) to overcome this limitation. PINROD extends dynamics-aware implicit neural representations~\cite{yin2022continuous} with parameterized neural ordinary differential equations, introducing parametric dependencies directly into the latent dynamics. This enables efficient and accurate modeling of nonlinear, high-dimensional oceanic data across diverse scenarios from a single trained model. Our experiments demonstrate that PINROD surpasses state-of-the-art methods in accuracy and computational efficiency, particularly in many-query settings.



\section{Methodology}

\subsection{Problem Setup}
Building on the need for efficient, generalizable ocean surrogate models, we aim to address the challenge of forecasting mesoscale dynamics under varying physical parameterizations. This is crucial for understanding ocean circulation sensitivity to parameter uncertainties, vital for robust climate projections. We focus on predicting the evolution of an idealized midlatitude ocean basin using the Simulating Ocean Mesoscale Activity (SOMA) test case within the Model for Prediction Across Scales-Ocean (MPAS-O) framework~\cite{deyoung2004challenges,ringler2013multi,petersen2019evaluation}. SOMA provides a computationally tractable yet realistic environment for evaluating a surrogate model. The SOMA configuration simulates an eddying ocean basin  (\(21.5^\circ\)N–\(48.5^\circ\)N, \(16.5^\circ\)W–\(16.5^\circ\)E) with curved coastlines, a continental shelf, and $32$ km horizontal resolution, capturing essential mesoscale dynamics. This setup is a more realistic evolution of typical double-gyre test cases, allowing for a better assessment of a forecasting surrogate's performance under conditions resembling real-world ocean modeling. We provide detailed data description in Appendix A.1. 

\textbf{Challenges.} Since the dataset contains prognostic variables, including temperature, salinity, and velocity fields, each defined over \(8521\) hexagonal cells with \(60\) vertical layers, it yields more than \(15\) million spatial and temporal data points per variable~\cite{wolfram2015diagnosing,sun2023surrogate}. Consequently, modeling this data presents challenges due to: 
\begin{enumerate}
    \setlength{\itemsep}{0pt}  
    \setlength{\parskip}{0pt}
    \item \textbf{High Dimensionality}: The fine spatial resolution and long temporal extent result in a high-dimensional dataset, posing significant computational challenges.
    \item \textbf{Unstructured Grid}: Compared to Cartesian grids, the hexagonal grid complicates interpolation and generalization.
    \item \textbf{Nonlinear and Parameter-Dependent Dynamics}: Ocean dynamics involve complex nonlinear interactions influenced by physical parameters like the Bottom Drag Coefficient \(C_d\), which controls near-bottom friction, affecting deep ocean currents and energy dissipation. These parameterizations introduce further variability and complexity and is important for parameter sensitivity analysis and uncertainty quantification.
    
\end{enumerate}

\textbf{Goal.} Our objective is to solve the initial value problem for parameterized prediction. Specifically, given the initial state of the ocean and a set of physical parameters, the goal is to predict the \textit{continuous spatiotemporal evolution} of these prognostic variables at arbitrary future times. This problem setting mimics real-world scenarios where ocean models are initialized with observed data and integrated forward in time to produce forecasts.

\textbf{Problem Formulation.} We formulate the SOMA test case prediction task as a parameterized system of PDEs with an initial condition:
\[
\partial_t u = F(t, x, u; \mu) \quad \text{with} \quad (t, x) \in [0, T] \times \mathbb{X},
\]

where \( u = u(t, x) \) represents the ocean state, $\mu \in \mathbb{R}^d$ denotes physical parameters, $F$ governs the nonlinear dynamics, and $\mathbb{X} \in \mathbb{R}^3$ is the spatial domain defined by SOMA's unstructured mesh. The initial condition is:
\[
u(0, x) = u^0(x) \quad \text{with} \quad x \in \mathbb{X},
\]
where \( u^0(x) \) specifies the initial ocean state. Given the intial state and physical paramter, the task involves forecasting $u(t, x)$ at arbitrary continuous spatial coordinates and 

\subsection{Model}

To address the challenge of the SOMA prediction task, we introduce \textbf{P}arameterized \textbf{I}mplicit \textbf{N}eural \textbf{R}epresentations for \textbf{O}cean \textbf{D}ynamics (\textbf{PINROD}), a reduced-order modeling framework integrating INRs and parametrized neural ODEs. The \textbf{PINROD} reduces the high-dimensionality of the data by encoding the spatial states into a low-dimensional latent vector, while the latent dynamics, influenced by physical parameters, are modeled using neural ODEs. This method learns the dynamic system's flow in a continuous manner, allowing it to be trained on the unstructured mesh of the SOMA data, while enabling prediction at continuous time and spatial positions. Overall, the framework comprises the three major components:

\textbf{Encoder.} The encoder $E_{\phi}$ maps the high-dimensional states $u_t \in \mathbb{R}^{N_t}$ to low-dimensional latent vectors $\alpha_t \in \mathbb{R}^k$ with $k \ll N_t$ as $\alpha_t = E_\phi(u_t) $.  
We employ an auto-decoder \cite{park2019deepsdf} to obtain the latent vectors $\alpha_t$ by minimizing the decoding error.  Given the decoder $D_{\phi}$, $\alpha_t$ is obtained by minimize the decoding loss $\| D_\phi(\alpha_t) - u_t\|^2_2$. 

\textbf{PNODE.} To capture the continuous temporal evolution of these latent states, we extend Neural ODEs to include explicit parametric dependencies. Specifically, we define:
\(
    \frac{d\alpha_t}{dt} = f_\theta(\alpha_t, \mu),
\)
where $\mu$ represents physical parameters. We embed $\mu$ via an MLP and concatenate it with $\alpha_t$ as the input to another MLP which outputs the predicted derivatives. This allows the learned dynamics to adapt to varying physical conditions.

\textbf{Decoder.} Once the latent states are evolved in time, we reconstruct the original high-dimensional fields using a coordinate-based INR decoder. In particular, we adopt a SIREN architecture, where the parameters of the SIREN network are modulated by the latent vector $\alpha_t$ as 
$ u_t(x) = D_{\theta_d}(\alpha_t)(x),$
where $x$ indexes the spatial coordinates. This allows us to produce predictions at any queried spatial location, including those not present in the training set, making the model highly flexible for unstructured meshes.

\textbf{Training Procedure.} We employ a two-step training strategy, in {Encoder--Decoder Pretraining:} We jointly optimize the encoder $E_\phi$, the decoder $D_\phi$, and the latent vectors $\{\alpha_t\}$ to minimize the reconstruction loss $\|D_\phi(\alpha_t) - u_t\|_2^2$. During this stage, the goal is to learn a low-dimensional manifold that faithfully represents the high-dimensional climate fields.
{PNODE Training:} We fix the encoder--decoder parameters (and hence the latent vectors) and train the PNODE to capture the continuous-time dynamics. Concretely, we optimize $f_\theta$ by minimizing the temporal prediction error on the latent trajectories, with physical parameters $\mu$ included as additional inputs.

\begin{wrapfigure}{r}{0.48\textwidth} 
    
    \centering
    \vspace{-15pt}
    \includegraphics[width=\linewidth]{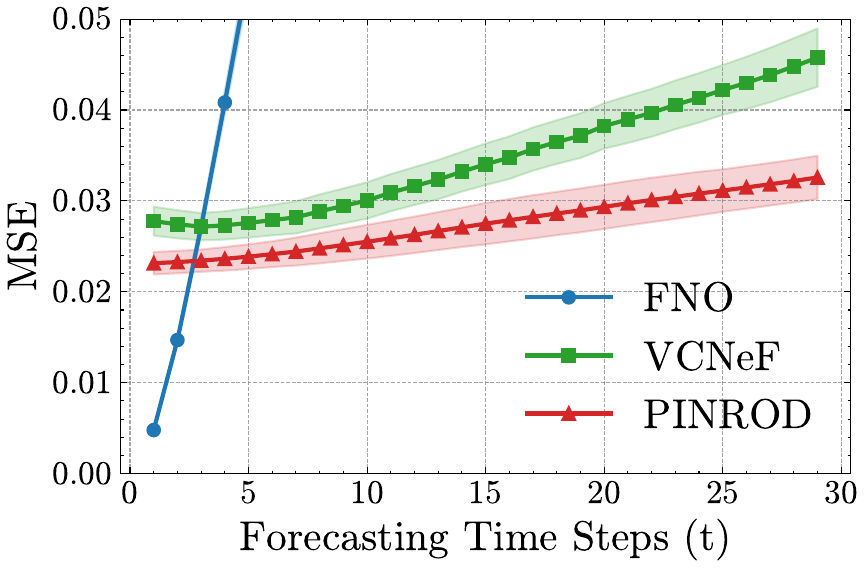}
    \caption{The mean squared error (MSE) of temperature predictions over 29 days. FNO exhibits accumulating error due to its autoregressive strategy, while VCNeF and PINROD maintain more stable performance.}
    \label{fig:error_over_time}
    \vspace{-20pt}
\end{wrapfigure}

This approach allows \textbf{PINROD} to handle unstructured sampling mesh, while incorporating relevant physical knowledge into the latent-state evolution. The result is a flexible framework that can make continuous forecasts on the SOMA dataset, surpassing grid-based baselines that struggle with unstructured data.

\section{Experiments}

We demonstrate the performance of \textbf{PINROD} on the SOMA dataset, comparing with two baselines FNO and VCNeF. We utilize SOMA simulations with varied bottom drag coefficient values generated from forward runs. The dataset contained 92 forward runs, each spanning 30 days of ocean states. 
Our prediction task involves that based on the initial state (day 0) and the bottom drag coefficient value, forecast the ocean states for the next 29 days (days 1-29).
We split 92 runs into 72 training set, 10 validation set and 10 testing set. The model performance is reported below. 

\textbf{Baselines.}  \textbf{(1)} Fourier Neural Operator (FNO) ~\cite{li2020fourier}, FNO is a neural operator method that learns function mappings in the Fourier frequency domain. In our setup,  FNO is trained to make one-day-ahead predictions and is tested to perform multi-days forecasting autoregressively. \textbf{(2)} Vectorized Conditional Neural Field (VCNeF) ~\cite{hagnberger2024vectorized}, VCNeF is a Vision Transformer (ViT) based conditional neural field. It takes physical parameters as condition and learns to solve initial value problem in continuous time. However, it requires training data on a regular grid. \textbf{(3)} PINROD. Our proposed method, which learns spatio-temporal dynamics on unstructured meshes.

\textbf{Metrics.} We evaluate performance using Mean Square Error and Relative $L_2$ error: $\|u_\text{pred} - u_\text{true}\|_2/\|u_\text{true}\|_2$. Additionally, we report \textit{inference time} to compare computational efficiency.

\begin{table}[tbp]
    \caption{\textbf{Quantitative results on the SOMA test set.} 
    Metrics include mean squared error (MSE), relative $L_2$ error (Rel. Err.), and \textit{average inference time} (seconds) to predict all 29 days from a single initial condition. Bolded entries indicate the best results in each category.}
    \centering
    \begin{tabular}{lcccccccc}
        \toprule
        & \multicolumn{2}{c}{Temperature} & \multicolumn{2}{c}{Salinity} & \multicolumn{2}{c}{LayerThickness} & \multicolumn{1}{c}{Infer. Time} \\
        \cmidrule(lr){2-3} \cmidrule(lr){4-5} \cmidrule(lr){6-7} \cmidrule(lr){8-8}
        Method & MSE & Rel. Err. & MSE & Rel. Err. & MSE & Rel. Err. & Time (s) \\
        \midrule
        FNO & 3.80e-01 & 5.67e-02       & 9.11e-02 & 1.09e-02                       & 3.23e-01  & 1.29e-02                  & \textbf{1.08} \\
        VCNeF & 3.35e-02 & 1.55e-02     & 2.64e-03 & 1.57e-03                      & 8.8e-03 & 3.02e-03               & 1.13 \\
        PINROD & \textbf{2.74e-02} & \textbf{1.53e-02} &\textbf{2.38e-03}  &\textbf{1.49e-03}  & \textbf{7.98e-03} & \textbf{2.87e-03}  & 3.43 \\
        \bottomrule
    \end{tabular}
    \label{tab:performance_comparison}
\end{table}

\paragraph{Experimental Setup}
The SOMA data is sampled on a hexagonal mesh within a basin domain. 
FNO and VCNeF require regular grid data, 
so we interpolate values on hexagonal mesh onto 3D grid and pad 0's for out of bound coordinates. In contrast, PINROD directly uses coordinate and value pairs for training, avoiding any interpolation error.

\begin{wrapfigure}{r}{0.48\textwidth}
    \centering
    \vspace{-15pt}
    \includegraphics[width=\linewidth]{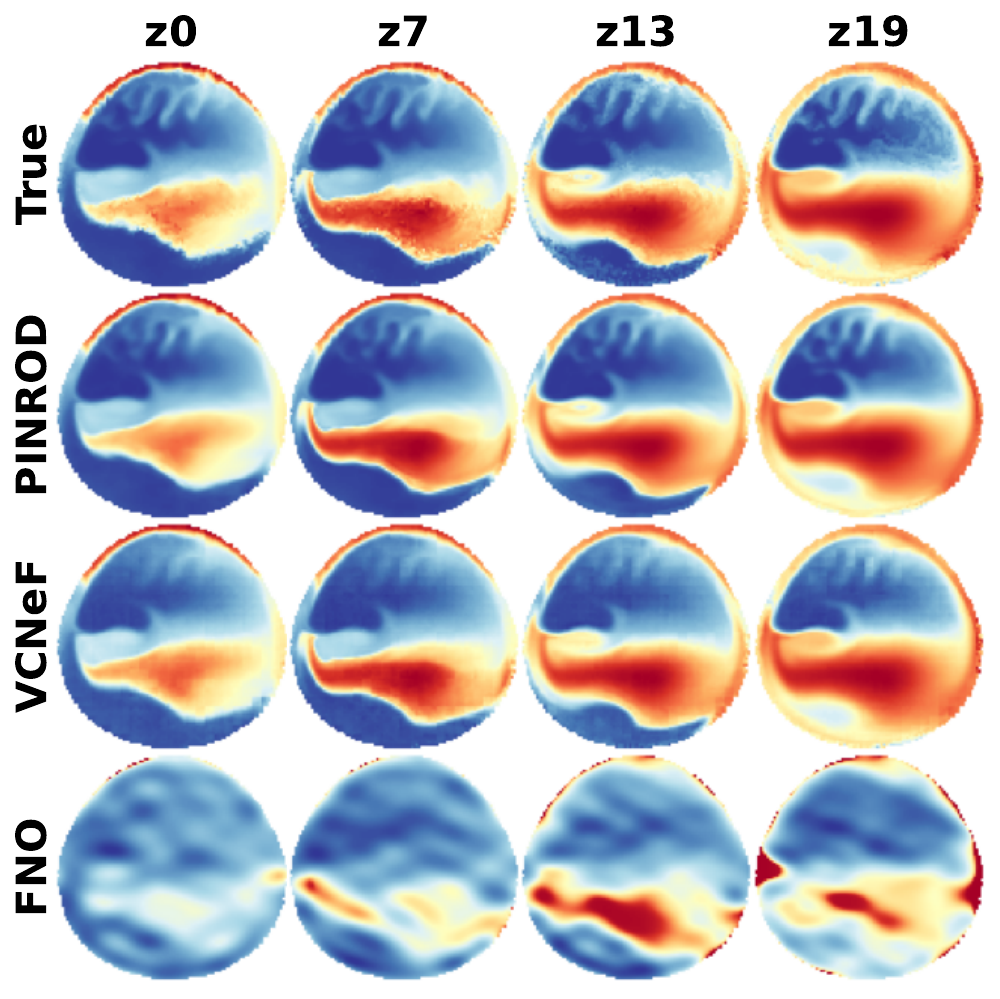}
    \caption{\textbf{Qualitative Comparison of Temperature Fields.}
    Snapshots at $t=29$ for different depths ($z=0,7,13,19$). PINROD more accurately preserves fine-scale structures.}
    \label{fig:image_matrix}
    \vspace{-15pt}
\end{wrapfigure}

\paragraph{Results and Discussion.}
Table \ref{tab:performance_comparison} provides quantitative results for three representative fields: Temperature, Layer Thickness and Salinity. From the table, we observe that PINROD consistently yields both lower MSE and lower relative $L_2$ error, reflecting PINROD's inherent advantage in handling unstructured dataset without interpolation. While PINROD's inference time is not the fastest among comparing methods, it remains significantly faster than numerical simulations.

Figure \ref{fig:error_over_time} illustrates how the temperature MSE over the 29-day forecast horizon. While FNO shows reasonable short term accuracy, its error accumulates rapidly due to autoregressive predictions. By contrast, the other two methods, VCNeF and PINROD maintain more stable errors over longer time spans. Figure \ref{fig:image_matrix} presents qualitative results for temperature prediction at $t=29$ across four different depth ( $z = $ 0, 7, 13, 19). The figures show that PINROD preserves fine-scale structures more accurately, which is crucial in climate modeling. We show more visualizations at $t=1, 9, 19$ in the Appendix A.2.


\section{Conclusion}
In this paper, we introduced PINROD, a novel method that combines implicit neural representations with parameterized nerual ODE for baroclinic ocean forecasting, we applied it to the challenging SOMA dataset. The method addresses key challenge of SOMA dataset, including unstructured sampling, high dimentionality and parammetric dependent non-linear dynamics.  Our experiments demonstrate that PINROD addresses these difficulites and outperforms compteting baselines. Our method advances the machine learning application in climate science, facilitate data-driven modeling of complex spatio-temporal processes in climate science problems. Future work can focus on leveraging adaptive sampling strategies and efficient training techniques to improve the model efficiency, enabling its application  to  more complex and large-scale oceanic datasets. 

\section{Acknowledgement}
This material is based upon work supported by the U.S. Department of Energy (DOE), Office of Science, Office of Advanced Scientific Computing Research, and Office of Biological and Environmental Research, through the Scientific Discovery through Advanced Computing (SciDAC) program, under Award Number 89233218CNA000001 and Contract No. DE-AC02-06CH11357. Additional support was provided by the DOE through the Los Alamos National Laboratory, which is operated by Triad National Security, LLC, for the National Nuclear Security Administration of the U.S. Department of Energy (Contract No. 89233218CNA000001). Brookhaven National Laboratory is supported by the DOE Office of Science under Contract No. DE-SC0012704 and FWP No. CC122. This research also used resources from the National Energy Research Scientific Computing Center (NERSC), a DOE Office of Science User Facility located at LBNL.


\bibliography{iclr2025_conference}

\begin{thebibliography}{23}
\providecommand{\natexlab}[1]{#1}
\providecommand{\url}[1]{\texttt{#1}}
\expandafter\ifx\csname urlstyle\endcsname\relax
  \providecommand{\doi}[1]{doi: #1}\else
  \providecommand{\doi}{doi: \begingroup \urlstyle{rm}\Url}\fi

\bibitem[Almeida et~al.(2010)Almeida, Silvestre, and Pascoal]{almeida2010cooperative}
Joao Almeida, Carlos Silvestre, and Antonio Pascoal.
\newblock Cooperative control of multiple surface vessels in the presence of ocean currents and parametric model uncertainty.
\newblock \emph{International Journal of Robust and Nonlinear Control}, 20\penalty0 (14):\penalty0 1549--1565, 2010.

\bibitem[Berkooz et~al.(1993)Berkooz, Holmes, and Lumley]{berkooz1993proper}
Gal Berkooz, Philip Holmes, and John~L Lumley.
\newblock The proper orthogonal decomposition in the analysis of turbulent flows.
\newblock \emph{Annual review of fluid mechanics}, 25\penalty0 (1):\penalty0 539--575, 1993.

\bibitem[Berlinghieri et~al.(2023)Berlinghieri, Trippe, Burt, Giordano, Srinivasan, {\"O}zg{\"o}kmen, Xia, and Broderick]{berlinghieri2023gaussian}
Renato Berlinghieri, Brian~L Trippe, David~R Burt, Ryan~James Giordano, Kaushik Srinivasan, Tamay {\"O}zg{\"o}kmen, Junfei Xia, and Tamara Broderick.
\newblock Gaussian processes at the helm (holtz): A more fluid model for ocean currents.
\newblock In \emph{International Conference on Machine Learning}, pp.\  2113--2163. PMLR, 2023.

\bibitem[Deser et~al.(2010)Deser, Alexander, Xie, and Phillips]{deser2010sea}
Clara Deser, Michael~A Alexander, Shang-Ping Xie, and Adam~S Phillips.
\newblock Sea surface temperature variability: Patterns and mechanisms.
\newblock \emph{Annual review of marine science}, 2\penalty0 (1):\penalty0 115--143, 2010.

\bibitem[DeYoung et~al.(2004)DeYoung, Heath, Werner, Chai, Megrey, and Monfray]{deyoung2004challenges}
Brad DeYoung, Mike Heath, Francisco Werner, Fei Chai, Bernard Megrey, and Patrick Monfray.
\newblock Challenges of modeling ocean basin ecosystems.
\newblock \emph{Science}, 304\penalty0 (5676):\penalty0 1463--1466, 2004.

\bibitem[Edwards \& Marsh(2005)Edwards and Marsh]{edwards2005uncertainties}
Neil~R Edwards and Robert Marsh.
\newblock Uncertainties due to transport-parameter sensitivity in an efficient 3-d ocean-climate model.
\newblock \emph{Climate dynamics}, 24:\penalty0 415--433, 2005.

\bibitem[Hagnberger et~al.(2024)Hagnberger, Kalimuthu, Musekamp, and Niepert]{hagnberger2024vectorized}
Jan Hagnberger, Marimuthu Kalimuthu, Daniel Musekamp, and Mathias Niepert.
\newblock Vectorized conditional neural fields: A framework for solving time-dependent parametric partial differential equations.
\newblock \emph{arXiv preprint arXiv:2406.03919}, 2024.

\bibitem[Johnson et~al.(2023)Johnson, Febvre, Gorbunova, Metref, Ballarotta, Le~Sommer, et~al.]{johnson2023oceanbench}
J~Emmanuel Johnson, Quentin Febvre, Anastasiia Gorbunova, Sam Metref, Maxime Ballarotta, Julien Le~Sommer, et~al.
\newblock Oceanbench: The sea surface height edition.
\newblock \emph{Advances in Neural Information Processing Systems}, 36:\penalty0 78275--78295, 2023.

\bibitem[Kim et~al.(2023)Kim, Lee, Kim, Cho, and Han]{kim2023generalizable}
Chiheon Kim, Doyup Lee, Saehoon Kim, Minsu Cho, and Wook-Shin Han.
\newblock Generalizable implicit neural representations via instance pattern composers.
\newblock In \emph{Proceedings of the IEEE/CVF Conference on Computer Vision and Pattern Recognition}, pp.\  11808--11817, 2023.

\bibitem[Lee et~al.(2021)Lee, Tack, Lee, and Shin]{lee2021meta}
Jaeho Lee, Jihoon Tack, Namhoon Lee, and Jinwoo Shin.
\newblock Meta-learning sparse implicit neural representations.
\newblock \emph{Advances in Neural Information Processing Systems}, 34:\penalty0 11769--11780, 2021.

\bibitem[Li et~al.(2020)Li, Kovachki, Azizzadenesheli, Liu, Bhattacharya, Stuart, and Anandkumar]{li2020fourier}
Zongyi Li, Nikola Kovachki, Kamyar Azizzadenesheli, Burigede Liu, Kaushik Bhattacharya, Andrew Stuart, and Anima Anandkumar.
\newblock Fourier neural operator for parametric partial differential equations.
\newblock \emph{arXiv preprint arXiv:2010.08895}, 2020.

\bibitem[Lu et~al.(2024)Lu, Zhao, Han, Gan, Zhou, Zhou, Fu, Wang, Zhou, and Zhang]{lu2024oxygenerator}
Bin Lu, Ze~Zhao, Luyu Han, Xiaoying Gan, Yuntao Zhou, Lei Zhou, Luoyi Fu, Xinbing Wang, Chenghu Zhou, and Jing Zhang.
\newblock Oxygenerator: Reconstructing global ocean deoxygenation over a century with deep learning.
\newblock In \emph{Forty-first International Conference on Machine Learning}, 2024.
\newblock URL \url{https://openreview.net/forum?id=0HUInAsdoo}.

\bibitem[Luo et~al.(2022)Luo, Nadiga, Park, Ren, Xu, and Yoo]{luo2022bayesian}
Xihaier Luo, Balasubramanya~T Nadiga, Ji~Hwan Park, Yihui Ren, Wei Xu, and Shinjae Yoo.
\newblock A bayesian deep learning approach to near-term climate prediction.
\newblock \emph{Journal of Advances in Modeling Earth Systems}, 14\penalty0 (10):\penalty0 e2022MS003058, 2022.

\bibitem[Luo et~al.(2024)Luo, Xu, Nadiga, Ren, and Yoo]{luo2024continuous}
Xihaier Luo, Wei Xu, Balu Nadiga, Yihui Ren, and Shinjae Yoo.
\newblock Continuous field reconstruction from sparse observations with implicit neural networks.
\newblock In \emph{The Twelfth International Conference on Learning Representations}, 2024.
\newblock URL \url{https://openreview.net/forum?id=kuTZMZdCPZ}.

\bibitem[Park et~al.(2019)Park, Florence, Straub, Newcombe, and Lovegrove]{park2019deepsdf}
Jeong~Joon Park, Peter Florence, Julian Straub, Richard Newcombe, and Steven Lovegrove.
\newblock Deepsdf: Learning continuous signed distance functions for shape representation.
\newblock In \emph{Proceedings of the IEEE/CVF conference on computer vision and pattern recognition}, pp.\  165--174, 2019.

\bibitem[Petersen et~al.(2019)Petersen, Asay-Davis, Berres, Chen, Feige, Hoffman, Jacobsen, Jones, Maltrud, Price, et~al.]{petersen2019evaluation}
Mark~R Petersen, Xylar~S Asay-Davis, Anne~S Berres, Qingshan Chen, Nils Feige, Matthew~J Hoffman, Douglas~W Jacobsen, Philip~W Jones, Mathew~E Maltrud, Stephen~F Price, et~al.
\newblock An evaluation of the ocean and sea ice climate of e3sm using mpas and interannual core-ii forcing.
\newblock \emph{Journal of Advances in Modeling Earth Systems}, 11\penalty0 (5):\penalty0 1438--1458, 2019.

\bibitem[Ringler et~al.(2013)Ringler, Petersen, Higdon, Jacobsen, Jones, and Maltrud]{ringler2013multi}
Todd Ringler, Mark Petersen, Robert~L Higdon, Doug Jacobsen, Philip~W Jones, and Mathew Maltrud.
\newblock A multi-resolution approach to global ocean modeling.
\newblock \emph{Ocean Modelling}, 69:\penalty0 211--232, 2013.

\bibitem[Schmid(2010)]{schmid2010dynamic}
Peter~J Schmid.
\newblock Dynamic mode decomposition of numerical and experimental data.
\newblock \emph{Journal of fluid mechanics}, 656:\penalty0 5--28, 2010.

\bibitem[Sun et~al.(2023)Sun, Cucuzzella, Brus, Narayanan, Nadiga, Van~Roekel, H{\"u}ckelheim, and Madireddy]{sun2023surrogate}
Yixuan Sun, Elizabeth Cucuzzella, Steven Brus, Sri Hari~Krishna Narayanan, Balu Nadiga, Luke Van~Roekel, Jan H{\"u}ckelheim, and Sandeep Madireddy.
\newblock Surrogate neural networks to estimate parametric sensitivity of ocean models.
\newblock \emph{arXiv preprint arXiv:2311.08421}, 2023.

\bibitem[Vallis(2017)]{vallis2017atmospheric}
Geoffrey~K Vallis.
\newblock \emph{Atmospheric and oceanic fluid dynamics}.
\newblock Cambridge University Press, 2017.

\bibitem[Wolfram et~al.(2015)Wolfram, Ringler, Maltrud, Jacobsen, and Petersen]{wolfram2015diagnosing}
Phillip~J Wolfram, Todd~D Ringler, Mathew~E Maltrud, Douglas~W Jacobsen, and Mark~R Petersen.
\newblock Diagnosing isopycnal diffusivity in an eddying, idealized midlatitude ocean basin via lagrangian, in situ, global, high-performance particle tracking (light).
\newblock \emph{Journal of Physical Oceanography}, 45\penalty0 (8):\penalty0 2114--2133, 2015.

\bibitem[Xie et~al.(2022)Xie, Takikawa, Saito, Litany, Yan, Khan, Tombari, Tompkin, Sitzmann, and Sridhar]{xie2022neural}
Yiheng Xie, Towaki Takikawa, Shunsuke Saito, Or~Litany, Shiqin Yan, Numair Khan, Federico Tombari, James Tompkin, Vincent Sitzmann, and Srinath Sridhar.
\newblock Neural fields in visual computing and beyond.
\newblock In \emph{Computer Graphics Forum}, volume~41, pp.\  641--676. Wiley Online Library, 2022.

\bibitem[Yin et~al.(2022)Yin, Kirchmeyer, Franceschi, Rakotomamonjy, and Gallinari]{yin2022continuous}
Yuan Yin, Matthieu Kirchmeyer, Jean-Yves Franceschi, Alain Rakotomamonjy, and Patrick Gallinari.
\newblock Continuous pde dynamics forecasting with implicit neural representations.
\newblock \emph{arXiv preprint arXiv:2209.14855}, 2022.

\end{thebibliography}
\bibliographystyle{iclr2025_conference}

\newpage
\appendix
\section{Appendix}

In this appendix, we provide additional details about the SOMA dataset and visual examples of the predicted fields at different timesteps.

\subsection{SOMA Dataset}

The Simulating Ocean Mesoscale Activity (SOMA) dataset, as part of the MPAS-Ocean model, offers a sophisticated simulation of an eddying mid-latitude ocean basin designed to study mesoscale activity and its sensitivity to model parameters. The model domain spans latitudes from $21.58^\circ$N to 48.58°N and longitudes from 16.58°W to 16.58°E, featuring a circular basin with curved coastlines, a 150km wide 100m deep continuntal shelf, and a longitudinally constant wind stress forcing. The SOMA dataset is implemented in MPAS-Ocean, which uses an unstructured hexagonal mesh with the 32km resolution. The dataset consists of 8521 hexagonal cells, each extending through 60 vertical layers, leading to over 15 million spatical and temporal data points per simulation. 

SOMA tracks five key prognostic variables that describe ocean state evolution: layer thickness, salinity, temperature, zonal velocity, and meridional velocity. These variables are influenced by external physical parameters that introduce variability into the simulation. The four primary perturbed parameters include the bottom drag coefficient, which controls near-bottom friction; the Gent–McWilliams (GM) constant diffusivity, which governs eddy-induced advection; the Redi constant diffusivity, responsible for isopycnal mixing; and the background vertical mixing diffusivity, which regulates vertical turbulence. These parameters play a crucial role in mesoscale dynamics, making SOMA an essential dataset for parameter sensitivity analysis. 

Each SOMA simulation runs for three years, with the first two years serving as a spin-up period, followed by daily data recording in the third year. The dataset includes 100 ensemble runs per perturbed parameter, allowing for extensive evaluation of ocean model responses. By enabling controlled perturbations of key oceanic parameters, SOMA provides an invaluable resource for developing machine learning surrogates and studying the impact of parameter variations on ocean circulation. It has been used to train neural network models to predict ocean state evolution and compute adjoint sensitivities for parameter optimization. The dataset's combination of high spatial resolution, long simulation duration, and multiple perturbation parameters makes it a powerful tool for advancing climate modeling, uncertainty quantification, and data-driven ocean forecasting.

\subsection{More Qualitative Comparison of Temperature Field}
\begin{figure}[ht]
    \centering
    \includegraphics[width=0.9\textwidth]{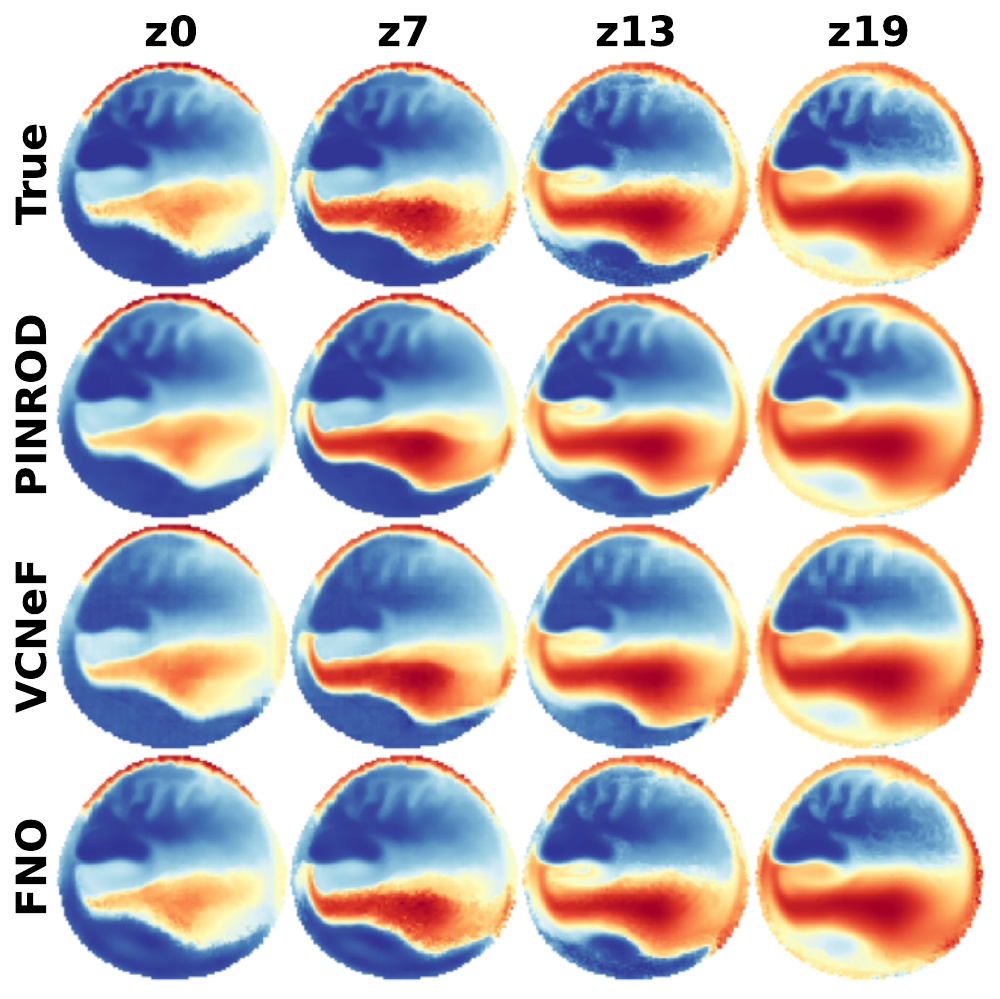}
    \caption{{Temperature field at \(\boldsymbol{t=0}\).} }
    \label{fig:appendix_t0}
\end{figure}

\begin{figure}[ht]
    \centering
    \includegraphics[width=0.9\textwidth]{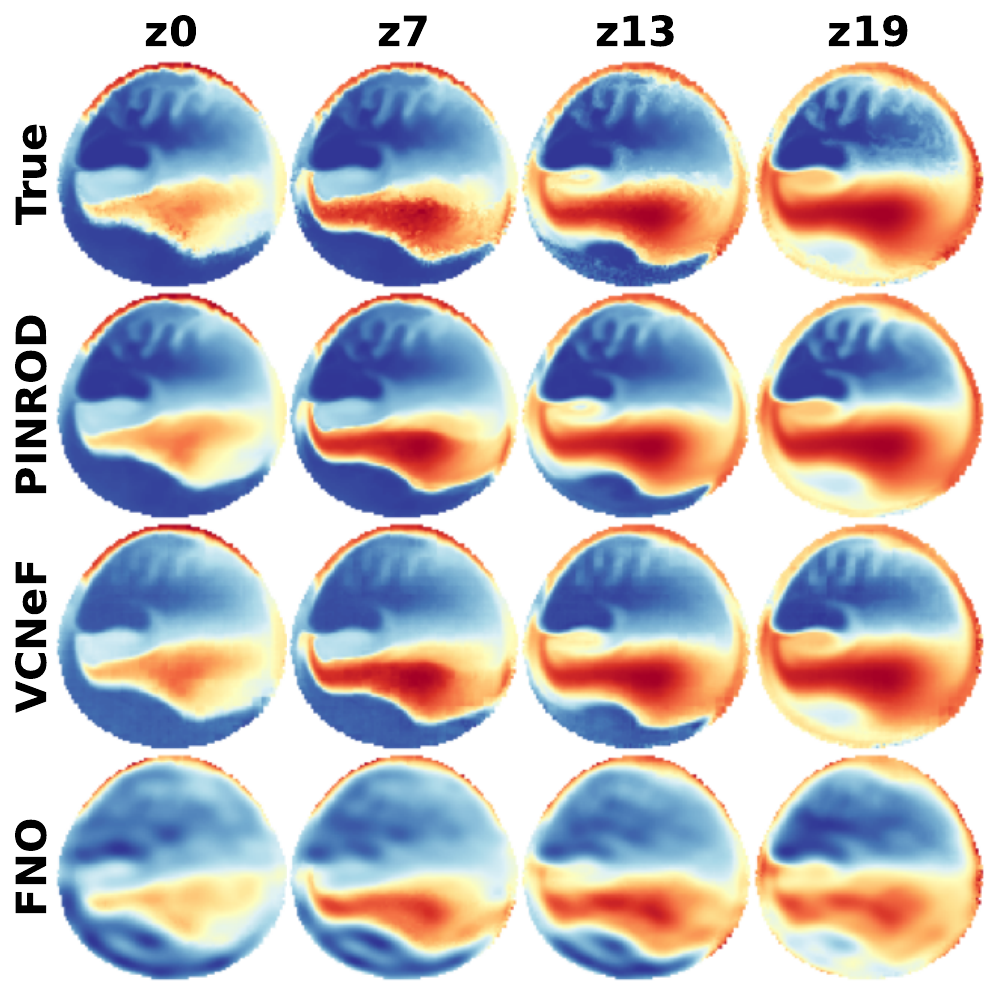}
    \caption{{Temperature field at \(\boldsymbol{t=9}\).}}
    \label{fig:appendix_t9}
\end{figure}

\begin{figure}[ht]
    \centering
    \includegraphics[width=0.9\textwidth]{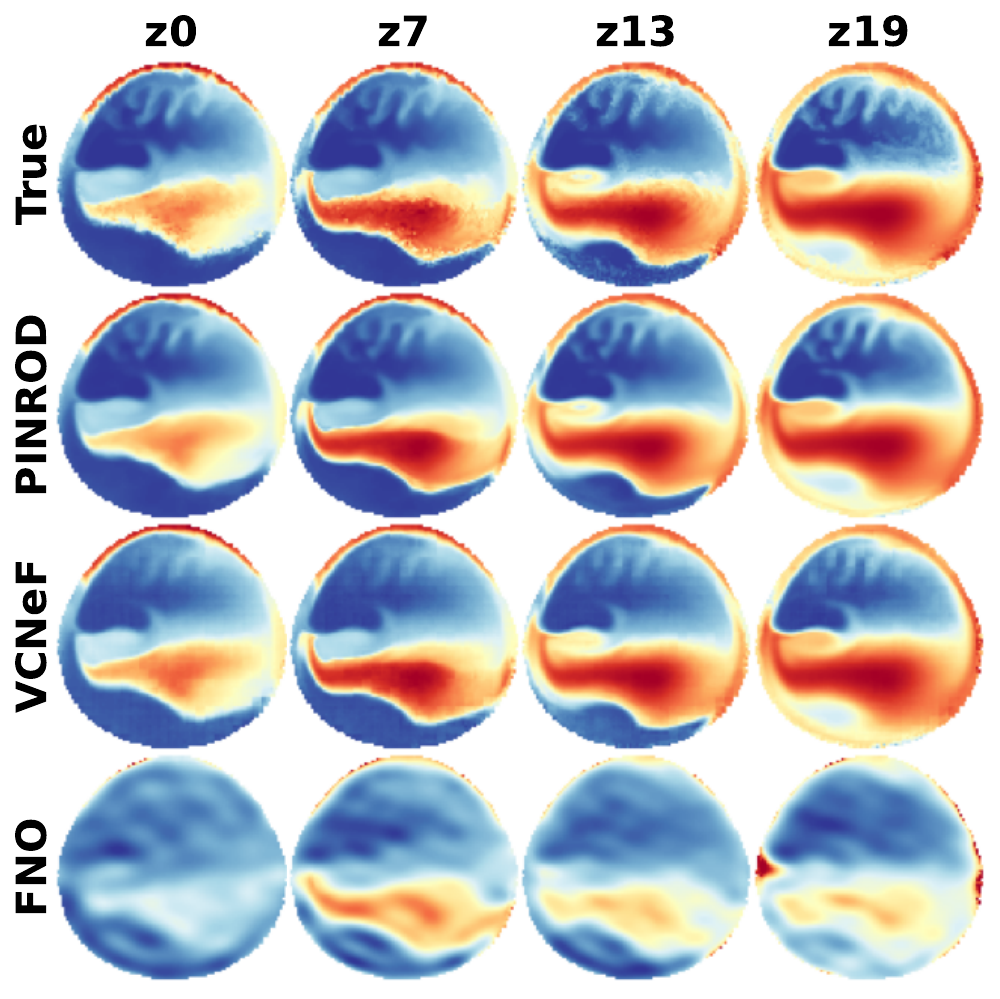}
    \caption{{Temperature field at \(\boldsymbol{t=19}\).}}
    \label{fig:appendix_t19}
\end{figure}

To complement the main text, Figures~\ref{fig:appendix_t0}, \ref{fig:appendix_t9}, and \ref{fig:appendix_t19} show temperature snapshots at \( t = 1, 9, \) and \( 19 \) for different depths (\(z = 0, 7, 13, 19\)). These examples illustrate how the predictions evolve over time, and how each method captures or misses fine-scale ocean state structures.

\end{document}